\documentclass[runningheads]{llncs}
\usepackage{graphicx}

\usepackage{amssymb}
\usepackage{amsmath}
\usepackage{wrapfig}
\usepackage{varwidth}
\usepackage{subcaption}
\begin{document}
%
\title{Knowledge-Based Sequential Decision-Making Under Uncertainty}
%
%

\author{Daoming Lyu \thanks{This report is for Doctoral Consortium session only.}}
\authorrunning{D. Lyu}
%

\institute{Auburn University, Auburn, AL 36849, USA \\
\email{daoming.lyu@auburn.edu}}
\maketitle              
\begin{abstract}
Deep reinforcement learning (DRL) algorithms have achieved great success on sequential decision-making problems, yet is criticized for the lack of data-efficiency and explainability. Especially, explainability of subtasks is critical in hierarchical decision-making since it enhances the transparency of black-box-style DRL methods and helps the RL practitioners to understand the high-level behavior of the system better. To improve the data-efficiency and explainability of DRL, declarative knowledge is introduced in this work and novel algorithm is proposed based on the symbolic representation. An experimental analysis on publicly available benchmarks validates the explainability of the subtasks and shows that our method can outperform the state-of-the-art approach in terms of data-efficiency.
\end{abstract}
\section{Introduction}
As shown in AlphaGo or autonomous driving, it is critical to making sequential decisions. Reinforcement Learning is a type of machine learning that is good at sequential decision-making problems. With the help of deep learning, deep reinforcement learning (DRL) algorithms have made a lot of achievements on sequential decision-making problems involving high-dimensional sensory inputs such as Atari games\cite{dqn:nature:2015}. However, there are some key issues in DRL approach, although it can learn policies that are able to surpass the overall performance of a professional human player. The first issue is the data inefficiency--DRL approach usually requires several millions of samples but still cannot learn long-horizon sequential actions for problems with sparse feedback and delayed rewards, such as Montezuma's Revenge \cite{dqn:nature:2015}. The second is about the explainability--the learning behavior based on the black-box neural network is nontransparent and hard to explain and understand. In real applications of decision-making, however, it is crucial to enable the system behavior to be explainable, in order to gain the confidence from the users and provide insights for their decision-making process \cite{gilpin2018explaining} with reasonable less data samples. To address the issues, we introduce the declarative knowledge into DRL and propose a framework of Symbolic Deep Reinforcement Learning (SDRL) by utilizing {\em Symbolic Planning} (SP) \cite{cim08}. In addition, we propose the {\em intrinsic goal}, a measurement of plan quality based on an internal utility function, to enable reward-driven planning. Therefore, the data efficiency in the decision-making process is improved from the meaningful exploration guided by symbolic planning \cite{yang:peorl:2018,lu2018robot}, and the explainability of the agent's behavior in task level can be achieved by the white-box algorithm of planning and reasoning with predefined and human-readable symbolic knowledge.
\section{SDRL Framework}
\begin{figure}
\centering
\includegraphics[height=3.5cm,width=8.cm]{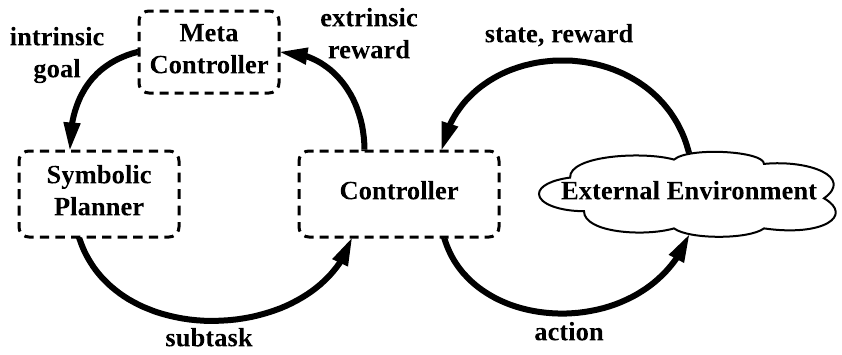}
\caption{Architecture of SDRL} \label{fig:arc}
\end{figure}

{\em Symbolic Deep Reinforcement Learning} (SDRL) framework features a {\em planner--controller--meta-controller} architecture, as shown in Fig.\ref{fig:arc}, which takes charge of subtask scheduling, data-driven subtask learning, and subtask evaluation, respectively.

We first assume a symbolic representation is given by human experts, which describes domain dynamics by causal rules. Note that the pre-defined symbolic representation is built for general-purpose. Next, we define the workflow as follows. 
The symbolic planner will generate high-level plans, i.e., a sequence of subtasks, to meet its {\em intrinsic goal}. An intrinsic goal is a measurement on plan quality, which approximates how much cumulative reward the plan may achieve. 
We assume a pre-trained mapping function can associate each sensory input with a symbolic state, i.e., performing symbol grounding, 
so that a set of options on the problem MDP can be induced based on symbolic states and the mapping function. We extend the reward structure of core MDP by introducing {\em intrinsic reward} and {\em extrinsic reward} to facilitate two levels of learning tasks. The sub-policies for the action level are learned using DRL algorithms based on intrinsic reward, with pseudo-rewards to encourage the agent to learn skills to achieve each subtask. As DRL continues, a metric is used to evaluate the competence of learned sub-policies, such as the success ratio over a number of episodes, from which extrinsic rewards is derived. When the sub-policy is learned and reliably achieves the subtask, the extrinsic reward is equivalent to the environmental reward. Using extrinsic rewards, meta-controller performs R-learning that reflects the long-term average reward and gains the reward of selecting each subtask. The learned values are returned to the symbolic planner and are used to measure plan quality and propose new intrinsic goals for the planner to improve the plan, by either exploring new subtasks or by sequencing learned subtasks that supposedly can achieve higher rewards in the next iteration.

In this process, the components of planner, controllers, and meta-controller cross-fertilize each other and eventually converge to an optimal symbolic plan along with the learned subtasks. While our framework is generic enough so that various planning and DRL techniques can be used, we instantiate our framework using action language ${\cal BC}$ for planning and R-learning for meta-controller learning.

\section{Experiment}
The proposed approach is evaluated on Taxi domain \cite{barto-sm:hrl} and Montezuma's Revenge~\cite{dqn:nature:2015}. Due to the space limitation, we skip the description of experimental settings and the complete results. Here we only show some results of Montezuma's Revenge. The interested readers are referred to  \cite{lyu2018sdrl} for more details.

 \begin{figure}[htb!]
 \centering
  \includegraphics[width=8.5cm, height=4.2cm]{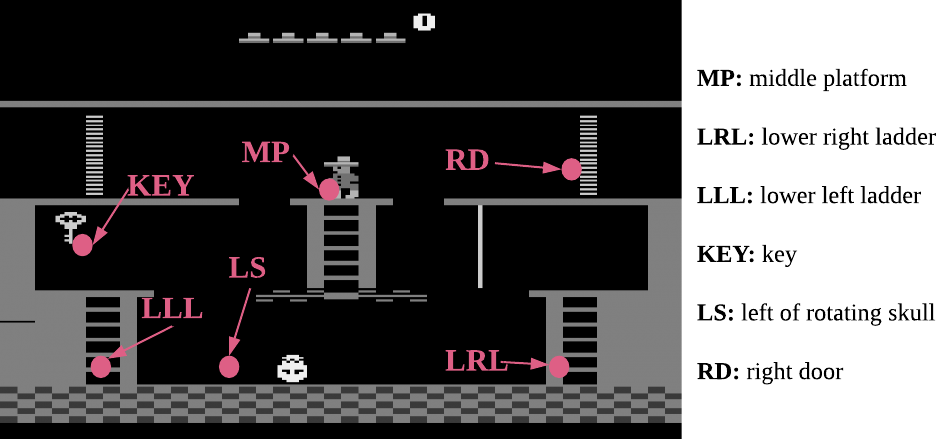}
 \caption{Pre-defined Locations or Objects}
 \label{fig:locations}
\end{figure}

\begin{figure}[htb!]
\centering
\begin{varwidth}{\linewidth}
\begin{verbatim}
% object declaration
location(mp;rd;ls;lll;lrl;key).
% dynamic causal law declaration
move(L) causes loc=L if location(L).
move(L) causes cost=L+Z if rho((at(L1)),move(L))=Z,
          loc=L1,picked(key)=false.
move(L) causes cost=L+Z if rho((at(L1),picked(key)),
          move(L))=Z,loc=L1,picked(key)=true.
inertial loc. inertial quality.
% static causal law declaration
picked(key)=true if loc=key.
nonexecutable move(key) if picked(key).
default rho((at(L1)),move(L))=10.     
default rho((at(L1),picked(key)),move(L))=10.
\end{verbatim}
\end{varwidth}
\caption{Montezuma's Revenge in $\mathcal{BC}$}
\label{kr}
\end{figure}

\begin{table}[htb!]
{\scriptsize
\begin{center}
\begin{tabular}{ c|c| c| c }
\hline\hline No.& subtask &  policy learned & in optimal plan  \\
  \hline
1&  MP to LRL, no key & \checkmark & \checkmark \\
2&  LRL to LLL, no key & \checkmark & \checkmark\\
3&  LLL to key, no key & \checkmark &\checkmark\\
4&  key to LLL, with key& \checkmark & \checkmark \\
5&  LLL to LRL, with or without key & \checkmark & \checkmark\\
6&  LRL to MP, with or without key & \checkmark & \checkmark \\
7&  MP to RD, with key & \checkmark & \checkmark\\
\hline
8&  LRL to LS, with or without key & \checkmark &  \\
9&  LS to key, with or without key & \checkmark &  \\
10& MP to RD, no key & \checkmark &  \\
\hline
11& LRL to key, with or without key &  &  \\
12 & key to LRL, with key &  &   \\
13 & LRL to RD, with key &  &  \\
\hline\hline
\end{tabular}
\end{center}}
\caption{Subtasks for Montezuma's Revenge}
 \label{tab:montezuma-subgoal}
\end{table}

\begin{figure}[htb!]
\caption{Experimental Results on Montezuma's Revenge}
\centering
\begin{subfigure}{.46\textwidth}
 \includegraphics[width=5.6cm, height=4.1cm]{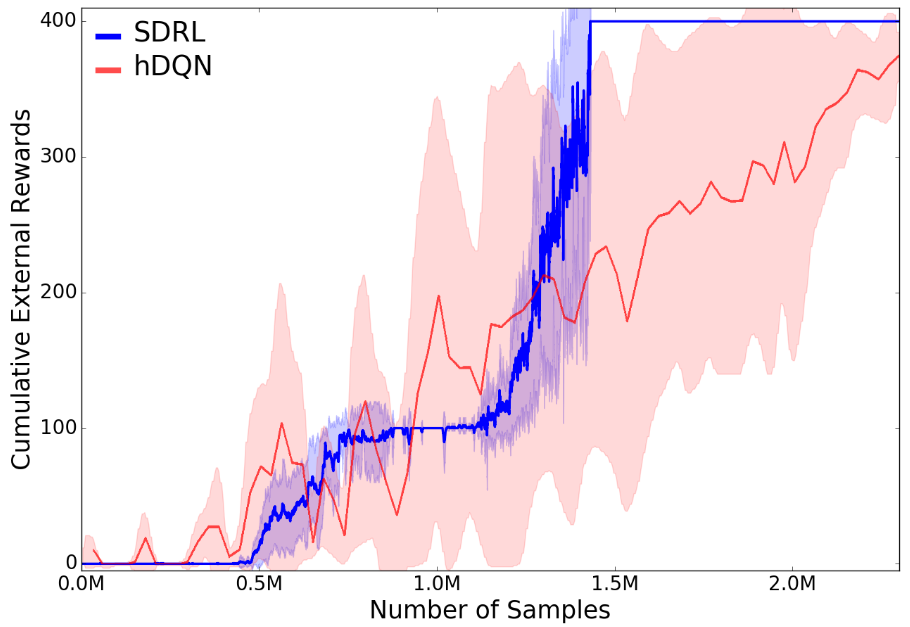}
   \caption{Learning Curve}
   \label{fig:cumulative}
\end{subfigure}
\quad
 \begin{subfigure}{.46\textwidth}
  \includegraphics[width=5.6cm, height=3.9cm]{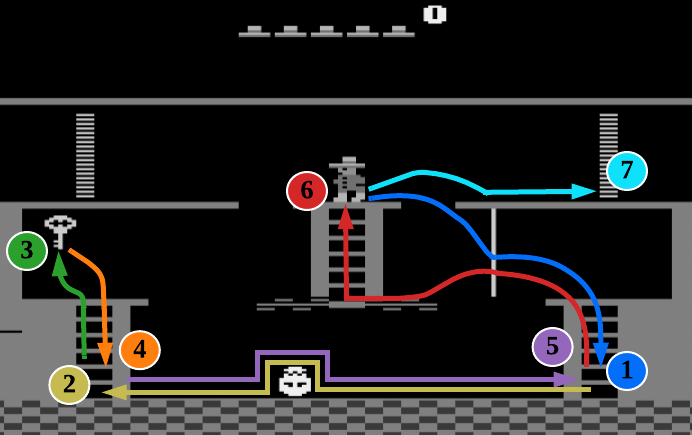}
 \caption{Final Solution}
 \label{fig:final}
\end{subfigure}
\end{figure}

As shown in Fig.\ref{fig:locations} and Fig.\ref{kr}, we formulated domain knowledge of Montezuma's Revenge in action language ${\cal BC}$ based on $6$ pre-defined locations or objects: middle platform ({\tt mp}), right door ({\tt rd}), left of rotating skull ({\tt ls}), lower left ladder ({\tt lll}), lower right ladder ({\tt lrl}), and key ({\tt key}). All 13 subtasks are pre-defined and shown in Table~\ref{tab:montezuma-subgoal}. Acutually, subtasks 1--10 can be successfully learned during the experiments, but only 7 of them (1--7) were selected in the final solution. This can be explained by the extrinsic rewards derived from training performance. For subtasks 11 -- 13, they were shown to be too difficult to learn in our experiments and discarded by the planner due to poor extrinsic rewards. In Fig.4 (Learning Curve), our method (SDRL) is compared with state-of-the-art approach, hierarchical DQN (hDQN) \cite{kulkarni2016deep}. The learning curve of SDRL shows that the agent first discovered the plan of collecting key after 0.5M samples by sequencing subtasks 1--3. Intrinsically motivated planning encourages exploring untried subtasks, and by learning more subtasks to move to other locations, the agent finally converges to the maximal cumulative external reward of $400$ around 1.5M samples by sequencing subtasks 1--7 (Fig.4 (Final Solution)). By comparison, hDQN cannot reliably achieve the score of $400$ around 2.5M samples. The shadow of the curves in Fig.4 (Learning Curve) represents the variance among multiple runs, which shows our SDRL has a small variance and can lead to more robust and stable learning.

\section{Conclusion}
This work demonstrates that by integrating symbolic planning with DRL for decision-making, explicitly represented symbolic knowledge can be used to perform high-level symbolic planning based on intrinsic goal which leads to improved task-level interpretability for DRL and data-efficiency.
This framework makes the final solution converge to an optimal symbolic plan along with the learned subtasks, bringing together the advantages of long-term planning capability with symbolic knowledge and end-to-end reinforcement learning. In the future work, one promising direction is to investigate on subtask discovery and we are working on this.

\bibliographystyle{splncs04}
\bibliography{reference}

\begin{thebibliography}{1}
\providecommand{\url}[1]{\texttt{#1}}
\providecommand{\urlprefix}{URL }
\providecommand{\doi}[1]{https://doi.org/#1}

\bibitem{barto-sm:hrl}
Barto, A., Mahadevan, S.: Recent advances in hierarchical reinforcement
  learning. Discrete Event Systems Journal  \textbf{13},  41--77 (2003)

\bibitem{cim08}
Cimatti, A., Pistore, M., Traverso, P.: Automated planning. In: van Harmelen,
  F., Lifschitz, V., Porter, B. (eds.) Handbook of Knowledge Representation.
  Elsevier (2008)

\bibitem{gilpin2018explaining}
Gilpin, L.H., Bau, D., Yuan, B.Z., Bajwa, A., Specter, M., Kagal, L.:
  Explaining explanations: An approach to evaluating interpretability of
  machine learning. arXiv preprint arXiv:1806.00069  (2018)

\bibitem{kulkarni2016deep}
Kulkarni, T.D., Narasimhan, K., Saeedi, A., Tenenbaum, J.: Hierarchical deep
  reinforcement learning: Integrating temporal abstraction and intrinsic
  motivation. In: Advances in Neural Information Processing Systems. pp.
  3675--3683 (2016)

\bibitem{lu2018robot}
Lu, K., Zhang, S., Stone, P., Chen, X.: Robot represention and reasoning with
  knowledge from reinforcement learning. arXiv preprint arXiv:1809.11074
  (2018)

\bibitem{lyu2018sdrl}
Lyu, D., Yang, F., Liu, B., Gustafson, S.: Sdrl: Interpretable and
  data-efficient deep reinforcement learning leveraging symbolic planning. In:
  AAAI (2019)

\bibitem{dqn:nature:2015}
Mnih, V., Kavukcuoglu, K., Silver, D., Rusu, A.A., Veness, J., Bellemare, M.G.,
  Graves, A., Riedmiller, M., Fidjeland, A.K., Ostrovski, G., et~al.:
  Human-level control through deep reinforcement learning. Nature
  \textbf{518}(7540),  529--533 (2015)

\bibitem{yang:peorl:2018}
Yang, F., Lyu, D., Liu, B., Gustafson, S.: Peorl: Integrating symbolic planning
  and hierarchical reinforcement learning for robust decision-making. In:
  International Joint Conference of Artificial Intelligence (IJCAI) (2018)

\end{thebibliography}

\end{document}